\documentclass[runningheads]{llncs}

\usepackage[T1]{fontenc}

\usepackage{multibib}
\newcites{languageresource}{Language Resources}
\usepackage{graphicx}
\usepackage{array}
\usepackage{multirow}
\usepackage{tabularx}
\usepackage{soul}
\usepackage{xspace}
\usepackage{todonotes}
\usepackage{enumitem}
\usepackage{subcaption}
\usepackage{epstopdf}
\usepackage[utf8]{inputenc}
\usepackage{caption}
\usepackage{hyperref}
\usepackage{xstring}
\usepackage{siunitx}
\usepackage{wrapfig}
\usepackage{color}
\usepackage{float}
\usepackage{caption}
\usepackage{booktabs}
\usepackage{newtxtext} 

\usepackage[utf8]{inputenc}
\usepackage{tcolorbox}
\usepackage{xcolor}
\usepackage{umoline}
\colorlet{red}{red!50!gray}    
\colorlet{blue}{blue!50!gray}    
\colorlet{orange2}{orange!40!gray} 
\colorlet{green}{green!30!gray} 

\newcommand\activity[1]{{\color{green} \Overline{{#1}}}}
\newcommand\activitydata[1]{{\color{orange} \Overline{{#1}}}}
\newcommand\furtherspecification[1]{{\color{purple} \Overline{{#1}}}}
\newcommand\gateway[1]{{\color{red} \Overline{{#1}}}}
\newcommand\conditionspecification[1]{{\color{orange2} \Overline{{#1}}}}
\newcommand\actor[1]{{\color{blue} \Overline{{#1}}}}

\newcommand{\pet}{process extraction from natural language text}

\newcommand{\PETD}{PET}

\begin{document}
\title{PET: An Annotated Dataset for Process Extraction from Natural Language Text Tasks}

\titlerunning{PET dataset}
\authorrunning{P. Bellan et al.}

\author{Patrizio Bellan\inst{1,2} \and 
		Han van der Aa\inst{3} \and 
		Mauro Dragoni\inst{1} \and \\ 
		Chiara Ghidini\inst{1} \and 
		Simone Paolo Ponzetto\inst{3} 
		}
\institute{
		Fondazione Bruno Kessler, Trento (Italy)  
		\and Free University of Bozen-Bolzano, Bolzano (Italy) \and University of Mannheim,  Mannheim (Germany) 
		\\ \email{\{pbellan\}@fbk.eu}
		}

\maketitle

\begin{abstract}
Process extraction from text is an important task of process discovery, for which various approaches have been developed in recent years.  
However, in contrast to other information extraction tasks, there is a lack of  gold-standard corpora of business process descriptions that are carefully annotated with all the entities and relationships of interest.
Due to this, it is currently hard to compare the results obtained by extraction approaches in an objective manner, whereas the lack of annotated texts also prevents the 
application of data-driven information extraction methodologies, typical of the natural language processing field. 
Therefore, to bridge this gap, we present the PET dataset, 
a first corpus of business process descriptions annotated with activities, gateways, actors, and flow information.  We present our new resource, including a variety of baselines to benchmark the difficulty and challenges of business process extraction from text.
	 PET can be accessed via \url{huggingface.co/datasets/patriziobellan/PET}

\keywords{Process Extraction from Text \and Business Process Management \and \\ Information Extraction, Natural Language Processing \and Dataset \and Gold Standard} 
\end{abstract}


\section{Introduction}
\label{sec:introduction}

Recent years have seen a growing interest of the Business Process Management (BPM) community on the task of extracting process models from text 
\cite{DBLP:conf/coling/ChallengesandOpportunitiesofApplyingNaturalLanguageProcessingBusinessProcessManagement,DBLP:conf/aiia/BellanDG20,Maqbool18}. 
Nonetheless, when investigated in light of modern data driven approaches for Information Extraction (IE), current work has major limitations~\cite{DBLP:conf/aiia/BellanDG20}. Arguably this is in part due to the limited availability of domain-specific, human annotated, gold-standard data that could be used to train from scratch or fine-tune data-driven methods, and which are essential to enable task-specific comparisons across competing approaches.
In fact, the availability of reference gold-standard datasets has the potential of fostering the application of NLP techniques to the  BPM field, and crucially makes clear what the applicability and limitations of state-of-the-art approaches for the domain of interest are.

With this work we aim to fill this gap and foster bridging of work in IE and data-driven BPM by providing a novel dataset of human-annotated processes in a corpus of process descriptions. The contributions of this work are:%
\begin{enumerate}[leftmargin=4mm, topsep=1pt]
    \item We provide a new reference corpus, annotation schema, and guidelines for the task of annotating business process models in running text. Our corpus includes annotations for different kinds of extraction levels, such as actors, activities and relations between them. As a basis for the annotated data, we used a collection of 45 textual descriptions, initially used by Friedrich et al.~\cite{friedrich2010master}.
    \item We quantify the difficulty of fundamental information extraction tasks for process model extraction by deploying a variety of baselines on our annotated data, thus providing an initial assessment of the feasibility of process extraction from natural language text. 
\end{enumerate}
Our vision builds upon bringing together heterogeneous communities such as NLP and BPM practitioners by defining shared tasks and resources (cf.\ previous work from \cite{nanni18} at the intersection of NLP and political science).
 All resources described in this paper (dataset, annotation guidelines, and inception schema) are freely available for the research community at \url{https://pdi.fbk.eu/pet-dataset}.

\section{Annotation Schema: Process Model Elements and Relations}
\label{sec:annotation-guidelines}
\begin{wrapfigure}{r}{.5\textwidth}
	\vspace{-1em}
    \includegraphics[width=.5\textwidth]{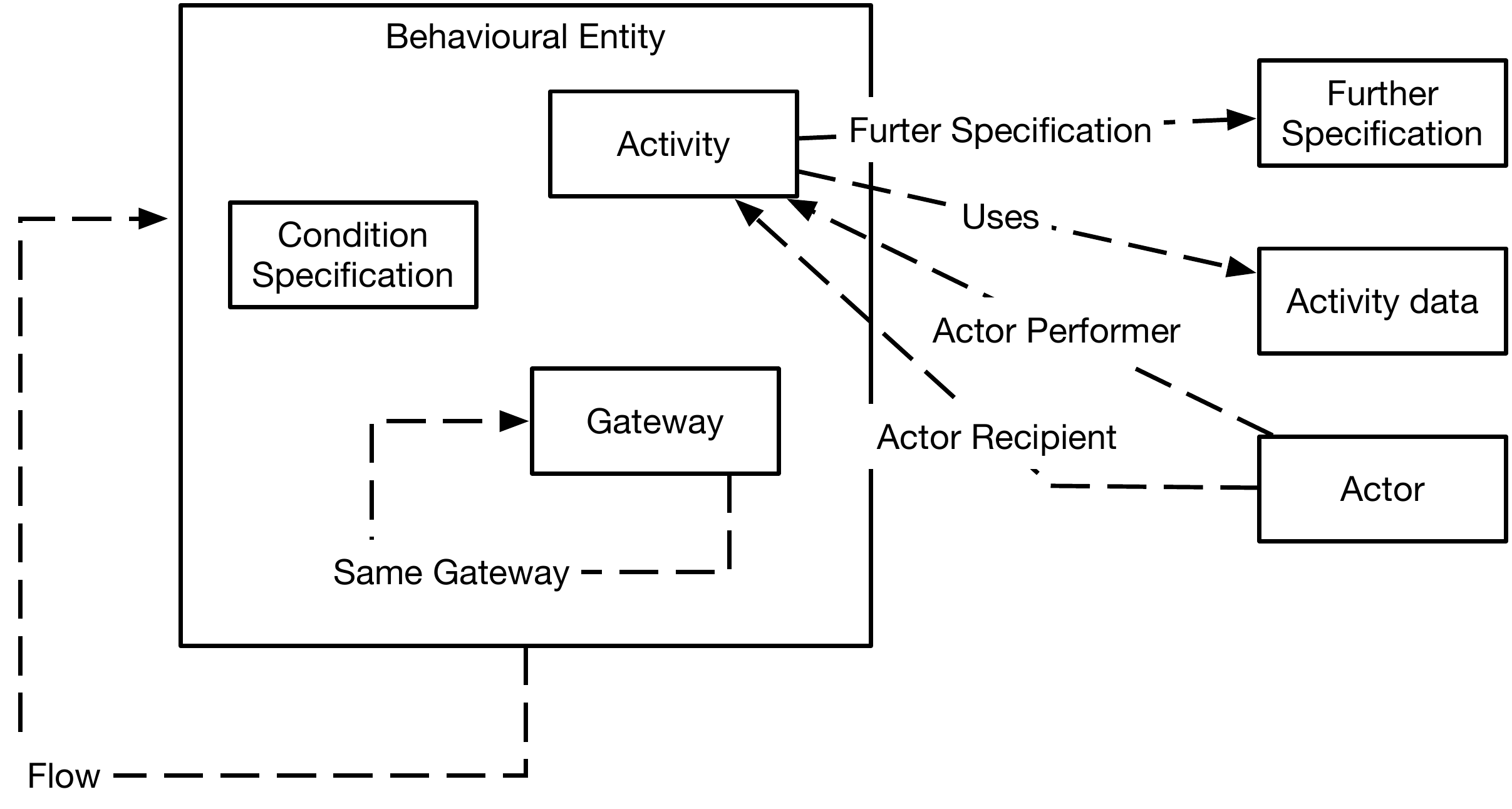}
  \caption{Annotated elements and relations.}
  \label{fig:annotationschema}
  	\vspace{-1em}
\end{wrapfigure}
In this section, we present the annotation schema we used to annotate textual process descriptions. For brevity, we here focus on the elements and their relations depicted in \autoref{fig:annotationschema}, whereas we refer the reader to the aforementioned URL for a 
complete description of the process elements, relations, annotation schema, and employed annotation rules.

\autoref{img:text-example} presents an excerpt of a procedural text with the respective process elements annotated. For brevity, we here omit the visualization of relations.
An \textit{Activity} represents a single task performed within a process (e.g., \activity{sends}),
while an associated \textit{Activity Data} element captures the object that an activity acts on, (e.g., \activitydata{the questionare})\footnote{Differently from customary BPM terminology, we decided to break down activity to differentiate among the activity ``action'' expression and the object the activity acts as NLP to ease the annotation task of different actions on the same object (or vice versa) and because NLP techniques may differ concerning the dealing of verb expressions and noun expressions.}.
These two elements are linked by the \textit{Uses} relation (e.g., \activity{sends}$\rightarrow$\activitydata{the questionnaire}). 
An \textit{Actor} defines the process participant involved in an activity execution (e.g., \actor{the customer office}). We differentiate between actors that perform/are responsible for an activity using the \textit{Actor Performer} relation (e.g., \activity{sends}$\rightarrow$\actor{the customer office}); and actors that receive the results of an activity using the \textit{Actor Recipient} relation (e.g., \activity{sends}$\rightarrow$\actor{the claimant}).
A \textit{Further Specification} element captures additional details of an activity that are relevant, but not covered by the other annotations, such as the means used to perform an activity (e.g., \furtherspecification{by email}).
An activity is connected to its further specification details by the relation of the same name (e.g., \activity{sends}$\rightarrow$\furtherspecification{by email}).

\begin{figure}[!btp]
	\centering
	\scalebox{0.9}
	{\begin{tcolorbox}
			\actor{The customer office} \activity{sends} \activitydata{the questionnaire} to \actor{the claimant} \furtherspecification{by email.} \\
			\gateway{If} \conditionspecification{the questionnaire is received}, \actor{the office} \activity{records} \activitydata{the questionnaire} and the process end.
			\\
			\gateway{Otherwise}, \activitydata{a reminder} is \activity{sent} to \actor{the customer}.
			
	\end{tcolorbox}}\
	\caption{
		{The example shows an example of a procedural text fragment with process elements annotated as follows:
			\small
			\textcolor{green}{{$\overline{Activity}$}}, \textcolor{orange}{{$\overline{Activity~ Data}$}}, \textcolor{blue}{{$\overline{Actor}$}}, \textcolor{purple}{{$\overline{Further~ Specification}$}}, \textcolor{red}{{$\overline{Gateway}$}}, \textcolor{orange2}{{$\overline{Condition~ Specification}$}}.
		}
	}
	\label{img:text-example}
\end{figure}

A \textit{Gateway} element represents a decision point (e.g., \gateway{if}), whereas the optional  \textit{condition specification} captures a prerequisite that a process execution instance must satisfy to enter a specific branch of a gateway (e.g., \conditionspecification{the questionnaire is received}). 
Currently, we cover two types of gateways: \textit{AND gateway} and \textit{XOR gateway}.
The \textit{Same Gateway} relation allows us to connect all the parts describing the same gateway, since its description may span over multiple sentences (e.g., \gateway{if}$\leftrightarrow$\gateway{otherwise}).
Finally, a \textit{Flow} relation defines the process logic by connecting all the activities, gateways, and condition specification elements in their flow sequence (e.g., \activity{sends}$\rightarrow$\gateway{if}). \emph{Flow} is thus also used to capture the relationship between a gateway and its condition(s) (e.g., \gateway{if}$\rightarrow$\conditionspecification{the questionnaire is received}).


\section{The PET Dataset}
\label{sec:dataset}
We used the annotation schema described in \autoref{sec:annotation-guidelines} and the associated annotation guidelines to establish the \PETD\ dataset described in this section.
This dataset is based on a set of textual descriptions initially used by Friedrich in their work on the extraction of process models from text~\cite{friedrich2010master}. 
The primary reason to use this data set as a basis is that the included 
textual descriptions are well-known within the community, which allows us to provide continuity to the investigation in this research area, as well as to start from a base set of textual documents that are in-line with the type of process narratives considered relevant by the community and used as a basis for the development of existing approaches.
However, no publicly available gold-standard annotation of the textual descriptions has been provided so far, a gap that we bridge with the \PETD\ dataset.

The \PETD\ dataset construction process has been split in five main phases:

\begin{enumerate}[topsep=1pt]
\item \textbf{Text pre-processing.} As the first operation, we check the content of each document and we tokenized it. 
	This initial check was necessary since the data were never validated.
	Indeed, several errors have been found and fixed.
\item \textbf{Text Annotation.}
 Each document has been assigned to three experts that were in change of identifying all the elements and flows with each document. In this phase, we used the Inception tool\footnote{\url{inception-project.github.io}} to support annotators. 
\item \textbf{Automatic annotation fixing.} After the second phase, we ran an automatic procedure relying on a rule-based script to automatically fix annotations that were not compliant with the guidelines.
\item \textbf{Agreement Computation.} Here, we computed, on the annotation provided by the experts, the agreement scores for each process element and for each relation between process elements pair adopting the methodology proposed in~\cite{DBLP:journals/jamia/HripcsakR05}.\footnote{We measured the agreement in terms of the F1 measure because, besides being straightforward to calculate, it is directly interpretable. Note that chance-corrected measures like $\kappa$ approach the F1-measure as the number of cases that raters agree are negative grows~\cite{DBLP:journals/jamia/HripcsakR05}.}
By following such a methodology, an annotation was considered in agreement among the experts if and only if they capture the same span of words and they assign the same process element tag to the annotation.
In the same way, a relation was considered in agreement if and only if the experts strictly annotated the same span of words representing (i) the process element related to the source element; (ii) the process element related to the target element; and, (iii) the relation tag between source and target.
The final agreement scores were obtained by averaging the individual scores obtained by the comparison of annotators pairs.
Table~\ref{tbl:agreements}
shows the annotation agreement computed for each process element and each process relation, respectively.
\item \textbf{Reconciliation.}
The last phase consisted of the mitigation of disagreements within the annotations provided by the experts. The aim of this phase is to obtain a shared and agreed set of gold standard annotations on each text for both entities and relations. Such entities also enable the generation of the related full-connected process model flow that can be rendered by using, but not limited to, a BPMN diagram.
During this last phase, among the 47 documents originally included into the dataset, 2 of them were discarded.
These texts were not fully annotated by the annotators since they were not be able to completely understand which process elements were actually included in some specific parts of the text, i.e., more than one interpretation would be provided.
\end{enumerate}

\begin{table}[!htb]
	\small
\begin{subtable}[c]{0.5\textwidth}
\centering
\begin{tabular}{lccc}
\toprule
& \textbf{Prec.}	& \textbf{Recall }& \textbf{F1} \\
\midrule
Activity & 0.96 & 0.87 & 0.91 \\
Activity Data & 0.93 & 0.73 & 0.82 \\
Actor & 0.96 & 0.84 & 0.89 \\
Further Specification & 0.43 &	0.33 &	0.37 \\
XOR Gateway	& 0.88	& 0.86	& 0.87	\\
AND Gateway	& 0.89	& 0.73	& 0.80	\\
Condition Specification	& 0.86	& 0.76	& 0.81	\\
\midrule
\textbf{Overall}	& 0.92 & 0.79 & 0.85 \\
\bottomrule
\end{tabular}
\end{subtable}%
\begin{subtable}[c]{0.5\textwidth}
\centering
\begin{tabular}{lccc}
\toprule
  & \textbf{Prec.} & \textbf{Recall} &	\textbf{F1} \\
\midrule
Sequence Flow & 1.00 & 0.67 & 0.80 \\
Uses & 1.00 & 0.72 & 0.83 \\
Actor Performer	& 1.00 & 0.74 & 0.85 \\
Actor Recipient	& 1.00 & 0.78 & 0.87 \\
Further Specification & 0.64 & 0.32 & 0.43 \\
Same Gateway & 0.88 & 0.72 & 0.80 \\
\midrule
\textbf{Overall}	& 0.98 & 0.69 & 0.81 \\
\bottomrule
\end{tabular}
\end{subtable}
\caption{Inter-annotator agreement for entities and relations.}
\label{tbl:agreements}
\vspace{-2em}
\end{table}

%
\noindent The final size of the dataset is 45 textual descriptions of the corresponding process models together with their annotations.
The total number of sentences of the dataset is 413, with an average sentences per document of 9.27 and each sentence has 18.15 words on average. 
%
%
Tables~\ref{tbl:entitiesstatistics} and~\ref{tbl:relationstatistics} contains the detailed statistic about process elements and relations respectively.

\begin{table*}[!htb]
\centering
\small
\begin{tabular}{lSSSSSSS}
\toprule

& {\multirow{2}{*}{\textbf{Activity}}}  &{\textbf{Activity}} &	{\multirow{2}{*}{\textbf{Actor}}}  &	{\textbf{Further}} & {\textbf{XOR}} &	{\textbf{AND}} & {\textbf{Condition}} \\
 	&  & {\textbf{data}} &  & {\textbf{specification}} & {\textbf{gateway}} & {\textbf{gateway}} & \textbf{specification} \\
\midrule
Absolute count & {501} & {452} & {439} & {64} & {117} & {8} & {80} \\
Relative count & 30.16\% & 27.21\% & 26.43\% & 3.86\% & 7.04\% & 0.48\% & 4.82\%	\\
Per document & 11.13 & 10.04 & 9.76 & 1.42 & 2.60 & 0.18 & 1.78 \\
Per sentence & 1.20 & 1.08 & 1.05 & 0.15 & 0.28 & 0.02 & 0.19 \\
Average length & 1.10 & 3.49 & 2.32 & 5.19 & 1.26 & 2.12 & 6.04 \\
Standard dev. & 0.48 & 2.47 & 1.11 & 3.40 & 0.77 & 1.54 & 3.04 \\
\bottomrule
\end{tabular}
\caption{Entity statistics.}
\label{tbl:entitiesstatistics}
\end{table*}%

\begin{table*}[!htb]
\small
\centering
\begin{tabular}{lSSSSSS}
\toprule
&  &  & {\textbf{Actor}} & {\textbf{Actor}}	& {\textbf{Further}} & {\textbf{Same}} \\
& {\textbf{Flows}} & {\textbf{Uses}}& {\textbf{Performer}} & {\textbf{Recipient}} & {\textbf{Specification}} & {\textbf{Gateway}} \\
\midrule
Absolute count	&	{689}	&	{477}	&	{313}	&	{164}	&	{64}	&	{43}	\\
Relative count	&	39.37\% &	27.26\%	&	17.89\%	&	9.37\%	&	3.66\%	&	2.46\%	\\
Count per document	&	15.31	&	10.60 	&   6.96	&	3.64	&	1.42	&	0.96	\\
Count per sentence	&	1.65	&	1.14	&	0.75	&	0.39	&	0.15	&	0.10	\\
\bottomrule
\end{tabular}
\caption{Relation statistics.}
\label{tbl:relationstatistics}
\vspace{-2em}
\end{table*}%

\section{Baseline Results}
\label{sec:baselines}
In this section, we present three baselines we developed to provide preliminary results obtained on the dataset and also to show how the dataset can be used to test different extraction approaches.
As described in Section~\ref{sec:dataset}, there are different type of elements that can be extracted (e.g., activities, actors, relations) and different assumptions that can be made (e.g., the exploitation of gold information or the process of the raw text).

From this perspective, we tested our baselines under three different settings and by using two different families of approaches: Conditional Random Fields (CRF) and Rule-Based (RB):

\begin{itemize}[topsep=1pt]
\item \textbf{Baseline 1 (B1)}: by starting from the raw text (i.e., no information related to process elements or relations has been used), a CRF-based approach has been used for building a model to support the extraction of single entities (e.g., activities, actors).
\item \textbf{Baseline 2 (B2)}: by starting from the existing gold information concerning the annotation of process elements, a RB strategy has been used for detecting relations between entities.
\item \textbf{Baseline 3 (B3)}: this baseline relies on the output of B1 concerning the annotations of process elements. Then, a RB strategy has been used for detecting relations between entities.
\end{itemize}

Concerning the CRF approach, we adopted a CRF model encoding data following the IOB2 schema.
Results have been obtained by performing a 5-folds cross-validation and by averaging observed performance.
While, concerning the RB approach, we defined a set of \textbf{six} rules taking into account text position of process elements.

Table~\ref{tbl:baselines} provides the results obtained by the three baseline approaches described above.
\begin{table*}
\begin{subtable}[c]{0.3\textwidth}
\centering
\scalebox{0.8}{
\begin{tabular}{lccc}
\toprule
& \multicolumn{3}{c}{\textbf{Baseline 1}} \\
\cmidrule{2-4}
& \textbf{Prec.}	& \textbf{Recall} & \textbf{F1} \\
\midrule
Activity & 0.91 & 0.73 & 0.81 \\
Activity Data & 0.87 & 0.58 &	0.70  \\
Actor & 0.90 & 0.67 &	0.76 \\
Further Specification & 0.38 &	0.13 &	0.19 \\
XOR Gateway	& 0.87	& 0.70	& 0.78	\\
AND Gateway	& 0.00	& 0.00	& 0.00	\\
Condition Specification	& 0.80 & 0.50 & 0.62 \\
\midrule
\textbf{Overall}	& 0.88 & 0.63 & 0.74 \\
\bottomrule
\end{tabular}
}
\end{subtable}%
\begin{subtable}[c]{0.9\textwidth}
\centering
\scalebox{0.8}{
\begin{tabular}{lcccccc}
\toprule
& \multicolumn{3}{c}{\textbf{Baseline 2}} & 
\multicolumn{3}{c}{\textbf{Baseline 3}} \\
\cmidrule(rl){2-4} \cmidrule(lr){5-7}
& \textbf{Prec.}	& \textbf{Recall} & \textbf{F1} & \textbf{Prec.}	& \textbf{Recall} & \textbf{F1} \\
\midrule
Sequence Flow & 1.00 & 0.787 &	0.88 & 1.00 & 0.37 &	 0.54 \\
Uses & 1.00 & 0.89 & 0.94 & 1.00 & 0.49 & 0.66 \\
Actor Performer	& 0.99 & 0.81 & 0.89 & 0.99 & 0.53 & 0.69 \\
Actor Recipient	& 0.99 & 0.82 & 0.90 & 1.00 & 0.48 & 0.65	\\
Further Specification & 1.00 & 0.83 & 0.91 & 0.88 & 0.11 & 0.19 \\
Same Gateway & 0.98 & 0.84 & 0.90 & 0.90 & 0.61 & 0.72	\\
\midrule
\textbf{Overall}	& 1.00 & 0.83 & 0.90 & 0.99 & 0.44 & 0.61 \\
\bottomrule
\end{tabular}
}
\end{subtable}
\caption{Baseline results for entity extraction and relation detection.}
\label{tbl:baselines}
\vspace{-3em}
\end{table*}%
An observation of baselines' performance highlights a general capability of the adopted approaches in detecting both process elements and relations with a high precision.
Exceptions are the \textit{further specification} and \textit{AND Gateway} elements, for which the baseline obtained very poor performance.
While on the one hand the observed precision is high, on the other hand the recall is the metric for which lower performance were obtained.
In turn, this affected the value of the F1 as well.
Hence, an interesting challenge worth of being investigated in this domain seems to be the detection of all elements rather then to detect them correctly.

\section{Conclusion}
\label{sec:conclusion}
In this paper, we presented the \PETD~dataset.
The dataset contains 45 documents containing narrative description of business process and their annotations.
Together with the dataset, we provided the set of guidelines we defined and adopted for annotating all documents.
The dataset building procedure has been described and, for completeness, we provided three baselines implementing straightforward approaches to give a starting point for designing the next generation of \pet~approaches.

\bibliographystyle{splncs04}
\bibliography{main}

\begin{thebibliography}{1}
\providecommand{\url}[1]{\texttt{#1}}
\providecommand{\urlprefix}{URL }
\providecommand{\doi}[1]{https://doi.org/#1}

\bibitem{DBLP:conf/coling/ChallengesandOpportunitiesofApplyingNaturalLanguageProcessingBusinessProcessManagement}
van~der Aa, H., Carmona, J., Leopold, H., Mendling, J., Padr{\'{o}}, L.:
  Challenges and opportunities of applying natural language processing in
  business process management. In: COLING. pp. 2791--2801 (2018)

\bibitem{DBLP:conf/aiia/BellanDG20}
Bellan, P., Dragoni, M., Ghidini, C.: A qualitative analysis of the state of
  the art in process extraction from text. In: Proc. of the AIxIA 2020
  Discussion Papers Workshop co-located with AIxIA2020. {CEUR} Workshop
  Proceedings, vol.~2776, pp. 19--30. CEUR-WS.org (2020)

\bibitem{friedrich2010master}
Friedrich, F.: Automated generation of business process models from natural
  language input. M. Sc., School of Business and Economics.
  Humboldt-Universit{\"a}t zu Berli  (2010)

\bibitem{DBLP:journals/jamia/HripcsakR05}
Hripcsak, G., Rothschild, A.S.: Technical brief: Agreement, the f-measure, and
  reliability in information retrieval. J. Am. Medical Informatics Assoc.
  \textbf{12}(3),  296--298 (2005)

\bibitem{Maqbool18}
Maqbool, B., Azam, F., Anwar, M.W., Butt, W.H., Zeb, J., Zafar, I., Nazir,
  A.K., Umair, Z.: A comprehensive investigation of {BPMN} models generation
  from textual requirements - techniques, tools and trends. In: Information
  Science and Applications. vol.~514, pp. 543--557. Springer (2018)

\bibitem{nanni18}
Nanni, F., Glavaš, G., Ponzetto, S.P., et~al.: Findings from the hackathon on
  understanding euroscepticism through the lens of textual data. In: LREC.
  European Language Resources Association (ELRA) (2018)

\end{thebibliography}
\end{document}